\documentclass[11pt,a4paper]{article}
\usepackage[hyperref]{naaclhlt2019}
\usepackage{times}
\usepackage{latexsym}

\usepackage{url}
\usepackage{graphicx}  %Required

\usepackage{times}
\usepackage{lipsum}
\usepackage{booktabs}
\usepackage{multirow}
\usepackage{latexsym} 
\usepackage{subcaption}
\usepackage[warn]{textcomp}
\usepackage{tabularx}
\usepackage{url}
\usepackage{enumitem}
\usepackage{diagbox}

\usepackage{amsmath}
\aclfinalcopy % Uncomment this line for the final submission
\def\aclpaperid{813} %  Enter the acl Paper ID here

\title{Evaluating Style Transfer for Text}

\author{Remi Mir$^1$, Bjarke Felbo$^2$, Nick Obradovich$^2$, Iyad Rahwan$^2$ \\
$^1$Department of EECS, Massachusetts Institute of Technology\\
$^2$Media Lab, Massachusetts Institute of Technology\\
}

\date{}

\begin{document}
\maketitle
\begin{abstract}
Research in the area of style transfer for text is currently bottlenecked by a lack of standard evaluation practices. This paper aims to alleviate this issue by experimentally identifying best practices with a Yelp sentiment dataset. We specify three aspects of interest (style transfer intensity, content preservation, and naturalness) and show how to obtain more reliable measures of them from human evaluation than in previous work. We propose a set of metrics for automated evaluation and demonstrate that they are more strongly correlated and in agreement with human judgment: direction-corrected Earth Mover's Distance, Word Mover's Distance on style-masked texts, and adversarial classification for the respective aspects. We also show that the three examined models exhibit tradeoffs between aspects of interest, demonstrating the importance of evaluating style transfer models at specific points of their tradeoff plots. We release software with our evaluation metrics to facilitate research.
\end{abstract} 

\section{Introduction}\label{intro}
Style transfer in text is the task of changing an attribute (style) of an input, while retaining non-attribute related content (referred to simply as \textit{content} for brevity in this paper).\footnote{This definition of style transfer makes a simplifying assumption that ``style" words can be distinguished from ``content" words, or words carrying relatively less or no stylistic weight, such as ``caf\`e" in ``What a nice caf\`e." The definition is motivated by penalizing unnecessary changes to content words, e.g. ``What a nice caf\`e" to ``This is an awful caf\`e."} For instance, previous work has modified text to make it more positive~\cite{Shen17}, romantic~\cite{StanfordDAR18}, or politically slanted~\cite{Prabhumoye}.

Some style transfer models enable modifications by manipulating latent representations of the text~\cite{Shen17,ZhaoKZRL17,fu2017style}, while others identify and replace style-related words directly~\cite{StanfordDAR18}. Regardless of approach, they are hard to compare as there is currently neither a standard set of evaluation practices, nor a clear definition of which exact aspects to evaluate. In Section~\ref{aspects}, we define three key aspects to consider. In Section~\ref{related_work}, we summarize issues with previously used metrics. Many rely on human ratings, which can be expensive and time-consuming to obtain.

To address these issues, in Section~\ref{methods}, we consider how to obtain more reliable measures of human judgment for aspects of interest, and automated methods more strongly correlated with human judgment than previously used methods. Lastly, in Section~\ref{results}, we show that the three examined models exhibit aspect tradeoffs, highlighting the importance of evaluating style transfer models at specific points of their tradeoff plots. We release software with our evaluation metrics at {\url{https://github.com/passeul/style-transfer-model-evaluation}.

\section{Aspects of Evaluation} \label{aspects}
We consider three aspects of interest on which to evaluate output text $x^\prime$ of a style transfer model, potentially with respect to input text $x$: 

\begin{enumerate}
  \item \textit{style transfer intensity} $STI(SC(x), SC(x^\prime))$ quantifies the difference in style, where $SC(\cdot)$ maps an input to a style distribution
  \item \textit{content preservation} $CP(x, x^\prime)$ quantifies the similarity in content between the input and the output
  \item \textit{naturalness} $NT(x^\prime)$ quantifies the degree to which the output appears as if it could have been written by humans
\end{enumerate}

Style transfer models should be compared across all three aspects to properly characterize differences. For instance, if a model transfers from negative to positive sentiment, but alters content such as place names, it preserves content poorly. 

\begin{table*}[]
\centering
\resizebox{\textwidth}{!}{
\begin{tabular}{|l|c|c|c|c|c|c|c|c|}
\hline
 & \multicolumn{3}{c|}{Style Transfer} & \multicolumn{3}{c|}{Content Preservation} & \multicolumn{2}{c|}{Naturalness} \\ \hline
 & HRC($x^\prime$) & HRD($x^\prime$) & SC($x^\prime$) & HRC($x$,$x^\prime$) & HRR($x$,{$\{x^\prime\}$})  & BLEU($x$,$x^\prime$) & HRC($x^\prime$) & PPL($x^\prime$) \\ \hline
CAAE &   & x & x &  & x  &   & \hspace{1.9mm}x$^\text{F}$ &   \\ \hline
ARAE &   & x & x & x &  & x & x & \hspace{1.9mm}x$^\text{F}$ \\ \hline
DAR  & x &   & x & x &  & x & \hspace{2.1mm}x$^\text{G}$ &   \\ \hline
\end{tabular}
}
\caption{Summary of past evaluation techniques. HRC is human rating on a continuous scale (e.g. 1 to 5). HRD is on discrete options (e.g. positive/negative). HRR is human ranking (most to least similar) of outputs, with respect to given input $x$. $\{x^\prime\}$ is the set of $x^\prime$ from models trained on different parameters. SC is a style classifier. PPL is perplexity. Superscripts denote that evaluation is done for fluency (F) or grammar (G), which we consider subsets of naturalness. Readers can see the original papers for details on methods falling under these techniques.}
\label{relatedwork_summary}
\end{table*}

\noindent If it preserves content well, but sequentially repeats words such as ``the", the output is unnatural. Conversely, a model that overemphasizes text reconstruction would yield high content preservation and possibly high naturalness, but little to no style transfer. All three aspects are thus critical to analyze in a system of style transfer evaluation.  

\section{Related Work}\label{related_work}

We review previously used approaches for evaluating the outputs of style transfer models. Due to the high costs related to obtaining human evaluations, we focus on three models: the cross-aligned autoencoder (CAAE), adversarially regularized autoencoder (ARAE), and delete-and-retrieve (DAR) models \cite{Shen17,ZhaoKZRL17,StanfordDAR18}. Table~\ref{relatedwork_summary} illustrates the spread of evaluation practices in these papers using our notation from Section~\ref{aspects}, showing that they all rely on a different combination of human and automated evaluation. For human evaluation, the papers use different instruction sets and scales, making it difficult to compare scores. Below we describe the automated metrics used for each aspect. Some rely on training external models on the corpus of input texts, $X$, and/or the corpus of output texts, $X^\prime$. We encourage readers seeking details on how to compute the metrics to reference the algorithms in the original papers.

\paragraph{Style Transfer}
Previous work has trained classifiers on $X$ and corresponding style labels, and measured the number of outputs classified as having a target style~\cite{Shen17,ZhaoKZRL17,StanfordDAR18}. Results from this \textit{target style scoring} approach may not be directly comparable across papers due to different classifiers used in evaluations.

\paragraph{Content Preservation}
To evaluate content preservation between $x$ and $x^\prime$, previous work has used BLEU~\cite{ZhaoKZRL17,StanfordDAR18}, an n-gram based metric originally designed to evaluate machine translation models~\cite{Papineni02}. BLEU does not take into account the aim of style transfer models, which is to alter style by necessarily changing words. Intended differences between $x$ and $x^\prime$ are thus penalized.

\paragraph{Naturalness}
Past evaluations of naturalness have relied largely on human ratings on a variety of scales under different
names: grammaticality, fluency/readability, and naturalness itself (Table \ref{relatedwork_summary}).
An issue with measuring grammaticality is that text with proper syntax can still be semantically nonsensical, e.g. ``Colorless green ideas sleep furiously"~\cite{ChomskySS}. Furthermore, input texts may not demonstrate perfect grammaticality or readability, despite being written by humans and thus being natural by definition (Section~\ref{aspects}).
This undermines the effectiveness of measures for such specific qualities of output texts.

\citet{ZhaoKZRL17} used perplexity to evaluate fluency, which, like grammaticality, we consider a subset of naturalness itself. Low perplexity signifies less uncertainty over which words can be used to continue a sequence, quantifying the ability of a language model to predict gold or reference texts \cite{Brown:1992,perplexityEqtn}. However, style transfer outputs are not necessarily gold standard, and the correlation between perplexity and human judgments of those outputs is unknown in the style transfer setting.

\section{Methods} \label{methods}

We describe how to construct a style lexicon for use in human and automated evaluations. We also describe best practices that we recommend for obtaining scores of those evaluations, as well as how they can be used for evaluating other datasets. Please refer to Section~\ref{results} for experimental results.

\begin{table} %[t!]
\begin{center}
\small{
\begin{tabular}{|c|c|}\hline 
Negative Sentiment & Positive Sentiment \\\hline
ruined & mouthwatering \\
worst & delightfully \\
failure & wonderfully \\
lackluster & marvelous \\
horrible & refreshing \\
\hline
\end{tabular}
}
\end{center}
\caption{Sample of words in a sentiment style lexicon.}
\label{sample_lex}
\end{table}

\subsection{Construction of Style Lexicon}\label{methods_style_lexicon}
Because the process of style transfer may result in the substitution or removal of more stylistically weighted words, it is ideal to have a lexicon of style-related words to reference. Words in $x$ and/or $x^\prime$ that also appear in the lexicon can be ignored in evaluations of content preservation.

While building a new style lexicon or an extension of existing ones like WordNet-Affect \cite{wordnet04} may be feasible with binary sentiment as the style, it may not be scalable to manually do so for various other types of styles. Static lexica also might not take context into account. This is an issue for text with words or phrases that are ambiguous in terms of stylistic weight, e.g. ``dog" in ``That is a man with a dog" vs. ``That man is a dog."

It is more appropriate to automate the construction of a style lexicon per dataset of interest. While multiple options may exist for doing so, we emphasize the simplicity and replicability of training a logistic regression classifier on $X$ and corresponding style labels. We populate the lexicon with features having the highest absolute weights, as those have the most impact on the outcome of the style labels. (Table \ref{sample_lex} shows sample words in the lexicon constructed for the dataset used in our experiments.) While sentiment datasets have been widely used in the literature \cite{Shen17,ZhaoKZRL17,StanfordDAR18}, a lexicon can be constructed for other datasets in the same manner, as long as the dataset has style labels.

Given existing NLP techniques, it may not be possible to correctly identify all style-related words in a text. Consequently, there is a tradeoff between identifying more style-related words and incorrectly marking some other (content) words as style-related. We opt for higher precision and lower recall to minimize the risk of removing content words, which are essential to evaluations of content preservation. This issue is not critical because researchers can compare their style transfer methods using our lexicon.

\subsection{Human Evaluation}
\label{sec:hum_eval}

As seen in Table~\ref{relatedwork_summary}, past evaluations of both style transfer and naturalness consider only output text $x^\prime$. Existing work from other fields have, however, shown that asking human raters to evaluate two relative comparisons provides more accurate scores than asking them to provide a numerical score for a single observation~\cite{stewart2005absolute,bijmolt1995effects}. With this knowledge, we construct more reliable ways of obtaining human evaluations via \textit{relative scoring} instead of \textit{absolute scoring}. 

\paragraph{Style Transfer Intensity}
Past evaluations have raters mark the degree to which $x^\prime$ exhibits a target style~\cite{StanfordDAR18}. We instead ask raters to score the difference in style between $x$ and $x^\prime$, on a scale of 1 (identical styles) to 5 (completely different styles). This approach can also used for non-binary cases. Consider text modeled as a distribution over multiple emotions (e.g. happy, sad, scared, etc.), where each emotion can be thought of as a style. One task could be to make a scared text more happy. Presented with $x$ and $x^\prime$, raters would still rate the degree to which they differ in style.

\paragraph{Content Preservation}
We consider the difficulty of asking raters to ignore style-related words as done in~\cite{Shen17}. Because not all raters may identify the same words as stylistic, their evaluations may vary substantially from one another. To account for this, we ask raters to evaluate content preservation on the same texts, but where we have masked style words using our style lexicon. Under this new ``masking" approach, raters have a simpler task, as they are no longer responsible for taking style into account when they rate the similarity of two texts on a scale of 1 to 5. 

\paragraph{Naturalness}
We ask raters to determine whether $x$ or $x^\prime$ (they are not told which is which) is more natural. An $x^\prime$ marked as more natural indicates some success on the part of the style transfer model, as it is able to fool the rater. This is in contrast to previous work, where raters score the naturalness of $x^\prime$ on a continuous scale without taking $x$ into account at all, even though $x$ serves as the basis for comparison of what is considered natural. 

\subsection{Automated Evaluation}\label{methods_autometrics}
In this section, we describe our approaches to automating the evaluation of each aspect of interest.

\paragraph{Style Transfer Intensity}
Rather than count how many output texts achieve a target style, we can capture more nuanced differences between the style distributions of $x$ and $x^\prime$, using Earth Mover's Distance~\cite{rubner1998metric,pele2009fast}. $EMD(SC(x), SC(x^\prime))$ is the minimum ``cost" to turn one distribution into the other, or how ``intense" the transfer is. Distributions can have any number of values (styles), so EMD handles binary and non-binary datasets. 

Note that even if $argmax(SC(x^\prime))$ is not the target style class, $EMD$ still acknowledges movement towards the target style with respect to $SC(x)$. However, we penalize (negate) the score if $SC(x^\prime)$ displays a relative change of style in the wrong direction, away from the target style.

Depending on $x$, not a lot of rewriting may be necessary to achieve a different style. This is not an issue, as $STI$ relies on a style classifier to quantify not the difference between the content of $x$ and $x^\prime$, but their style distributions. For the style classifier, we experiment with textcnn~\cite{kim2014convolutional,dongjun-textcnn} and fastText~\cite{JoulinGBM16}.

\paragraph{Content Preservation}

\begin{table}[]
\centering
\footnotesize{
\begin{tabular}{|l|}
\hline
\begin{tabular}[c]{@{}l@{}}\textbf{No modification}\\ Input: the girls up front \textit{incompetent} .\\ Output: the girls up front are \textit{amazing} .\end{tabular} \\ \hline
\begin{tabular}[c]{@{}l@{}}\textbf{Style removal}\\ Input: the girls up front .\\ Output: the girls up front are .\end{tabular}                      \\ \hline 
\begin{tabular}[c]{@{}l@{}}\textbf{Style masking}\\ Input: the girls up front \textit{$\langle customstyle\rangle$}.\\ Output: the girls up front are \textit{$\langle customstyle\rangle$}.\end{tabular}    \\ \hline
\end{tabular}
}
\caption{Text under different settings of style-based modification, as used in evaluations of content preservation. The sample output is from ARAE ($\lambda=1$).}
\label{style-correction}
\end{table}

We first subject texts to different settings of modification: style removal and style masking. This is to address undesired penalization of metrics on texts expected to demonstrate changes after style transfer (Section~\ref{related_work}). For style removal, we remove style words from $x$ and $x^\prime$ using the style lexicon. For masking, we replace those words with a $\langle customstyle\rangle$ placeholder. Table \ref{style-correction} exemplifies these modifications.

For measuring the degree of content preservation, in addition to the widely used BLEU, we consider METEOR and embedding-based metrics. METEOR is an n-gram based metric like BLEU, but handles sentence-level scoring more robustly, allowing it to be both a sentence-level and corpus-level metric \cite{meteor05}. 

For the embedding-based metrics, word embeddings can be obtained with methods like Word2Vec \cite{word2vec13} or GloVe \cite{glove14}. Sentence-level embeddings can be comprised of the most extreme values of word embeddings per dimension (\textit{vector extrema}) \cite{vecExtrema}, or word \textit{embedding averages} \cite{maluuba17}. \textit{Word Mover's Distance} (WMD), based on $EMD$, calculates the minimum ``distance" between word embeddings of $x$ and of $x^\prime$, where smaller distances signify higher similarity \cite{wmd15}. \textit{Greedy matching} greedily matches words in $x$ and $x^\prime$ based on their embeddings, calculates their similarity (e.g. cosine similarity), and averages all the similarities. It repeats the process in the reverse direction and takes the average of those two scores \cite{greedyMatch}.

We evaluate with all these metrics to identify the one most strongly correlated with human judgment of content preservation.

\paragraph{Naturalness}
For a baseline understanding of what is considered ``natural," any method used for automated evaluation of naturalness requires the human-sourced input texts. We train unigram and neural logistic regression classifiers~\cite{bowman} on samples of $X$ and $X^\prime$ for each transfer model. Via adversarial evaluation, these classifiers must distinguish human-generated inputs from machine-generated outputs. The more natural an output is, the  likelier it is to fool a classifier~\cite{jurafsky2014speech}. We calculate agreement between each type of human evaluation (Section \ref{sec:hum_eval}) and each classifier $AC$. Agreement is the ratio of instances where humans and $AC$ rate a text as more natural than the other.

We also train LSTM language models \cite{lstm} on $X$ and compute sentence-level perplexities for each text in $X^\prime$ in order to determine the relative effectiveness of adversarial classification as a metric.

\begin{table*}[]
\centering
\footnotesize{
\begin{tabular}{|lll|}\hline
\multicolumn{1}{|l|}{Input} & would \textit{n't recommend} until management works on friendliness and communication with residents .              &  \\  \hline
\multicolumn{1}{|l|}{ARAE ($\lambda=1$)}   & highly \textit{recommend} this place while living in tempe and management .                        &  \\ 
\multicolumn{1}{|l|}{CAAE ($\rho=0.5$)}   & would highly \textit{recommend} management on duty and staff on business .              &  \\ 
\multicolumn{1}{|l|}{DAR ($\gamma=500$)} & until management works on friendliness and is a \textit{great} place for communication with residents .             &\\ \hline
\end{tabular}
\caption{Sample outputs of a negative to positive sentiment style transfer task. Italicized words are style-related, according to a style lexicon. They can be masked or removed in evaluations of content preservation (Section \ref{methods_autometrics}).}
\label{style-transfer-model-output-samples}
}
\end{table*} 

\section{Experiments and Results}\label{results}
Due to high costs of human evaluation, we focus on CAAE, ARAE, and DAR models with transfer tasks based on samples from the Yelp binary sentiment dataset \cite{Shen17}.\footnote{Like most literature, including the papers on CAAE, ARAE and DAR, we focus on the binary case. Creating a high-quality, multi-label style transfer dataset for evaluation is a demanding task, which is out of scope for this paper.} Below we detail the range of parameters each model is trained on in order to compare evaluation practices and generate aspect tradeoff plots. Each of three Amazon Turk raters evaluated 244 texts per aspect, per model. Of those texts, half are originally of positive sentiment transferred to negative, and vice versa.

For brevity, we reference average scores (correlation, kappa, and agreement, each of which is described below) from across all models in our analysis of results. For detailed scores per model, please refer to the corresponding tables.

\subsection{Style Transfer Models}\label{model_details}
For each style transfer model, we choose a wide range of training parameters to allow for variation of content preservation, and indirectly, of style transfer intensity, in $X$. We show sample outputs from the models for a given input text in Table \ref{style-transfer-model-output-samples}.

CAAE uses autoencoders \cite{autoencoderDef} that are cross-aligned, assuming that texts already share a latent content distribution \cite{Shen17}. It uses latent states of the RNN and multiple discriminators to align distributions of texts in $X^\prime$ exhibiting one style with distributions of texts in $X$ exhibiting another. Adversarial components help separate style information from the latent space where inputs are represented. We train CAAE on various values (0.01, 0.1, 0.5, 1, 5) of $\rho$, a weight on the adversarial loss.

CAAE is a baseline for other style transfer models, such as ARAE, which trains a separate decoder per style class \cite{ZhaoKZRL17}. We train ARAE on various values (1, 5, 10) of $\lambda$, which is also a weight on adversarial loss.

The third model that we evaluate, which also uses CAAE as a baseline, avoids adversarial methods in an approach called Delete-and-Retrieve (DAR) \cite{StanfordDAR18}. It identifies and removes style words from texts, searches for related words pertaining to a new target style, and combines the de-stylized text with the search results using a neural model. We train DAR on $\gamma = 15$, where $\gamma$ is a threshold parameter for the maximum number of style words that can be removed from texts, with respect to the size of the corpus vocabulary. For this single training value, we experiment with a range of $\gamma$ values (0.1, 1, 15, 500) during test time because, by design, the model does not need to be retrained \cite{StanfordDAR18}.

\subsection{Human Evaluation}

\begin{table}[]
\centering
\footnotesize{
\begin{tabular}{|l|c|c|}
\hline
\multirow{2}{*}{Model} & \multicolumn{2}{c|}{Text Modification Setting} \\ \cline{2-3} 
                                & Unmasked        & Style Masked        \\ \hline
CAAE                            & 0.158          & \textbf{0.289}               \\ \hline
ARAE                            & 0.201          & \textbf{0.321}               \\ \hline
DAR                             & 0.161          & \textbf{0.281}               \\ \hline \hline
Average                             & 0.173          & \textbf{0.297}               \\ \hline
\end{tabular}
}
\caption{Fleiss' kappas for human judgments of content preservation of unmasked and style-masked texts.}
\label{cp_kappas}
\end{table}

\begin{table}[]
\centering
\footnotesize{
\begin{tabular}{|l|c|c|c|}
\hline
\multirow{2}{*}{Model} & \multicolumn{2}{c|}{Absolute}  & \multicolumn{1}{c|}{\multirow{2}{*}{Relative}} \\ \cline{2-3}
                                & $\tau=3$  & $\tau=2$ & \multicolumn{1}{c|}{}                          \\ \hline
CAAE                            & 0.193 & 0.321         & \textbf{0.579}                                          \\ \hline
ARAE                            & 0.215 & 0.415         & \textbf{0.741}                                          \\ \hline
DAR                             & 0.103 & 0.201         & \textbf{0.259}                                           \\ \hline \hline
Average                             & 0.170 & 0.312         & \textbf{0.526}                                           \\ \hline
\end{tabular}
} 
\caption{Fleiss' kappas for human judgments of absolute naturalness and relative naturalness of texts.}
\label{fleiss_nat_full}
\end{table}

\begin{table*}[]
\centering
\resizebox{\textwidth}{!}{%
\footnotesize{
\begin{tabular}{|l|c|c|c|c|}
\hline
\multirow{2}{*}{Model} & \multicolumn{2}{c|}{fastText}               & \multicolumn{2}{c|}{textcnn}               \\ \cline{2-5} 
                                & Target Style Scores & Earth Mover's Distance           & Target Style Scores & Earth Mover's Distance           \\ \hline \hline
CAAE                            & 0.566 $\pm$ 0.038               & \textbf{0.573 $\pm$ 0.038}  & 0.587 $\pm$ 0.037               & \textbf{0.589 $\pm$ 0.037} \\ \hline
ARAE                            & 0.513 $\pm$ 0.053               & \textbf{0.516 $\pm$ 0.053} & 0.515 $\pm$ 0.053               & \textbf{0.519 $\pm$ 0.053} \\ \hline
DAR                             & 0.470 $\pm$ 0.049               & \textbf{0.539 $\pm$ 0.045} & 0.508 $\pm$ 0.047               & \textbf{0.566 $\pm$ 0.043} \\ \hline \hline
Average                             & 0.516 $\pm$ 0.047               & \textbf{0.543 $\pm$ 0.045} & 0.537 $\pm$ 0.046               & \textbf{0.558 $\pm$ 0.044} \\ \hline
\end{tabular}
}
}
\caption{Correlations of automated style transfer intensity metrics with human scores.}
\label{sti-corr-full}
\end{table*}

% style removal and masking tables are kept separate
\begin{table*}[ht!]
\centering
\resizebox{\textwidth}{!}{%
\begin{tabular}{|l|c|c|c|c|c|c|}
\hline
Model & BLEU & METEOR & Embed Average & Greedy Match & Vector Extrema & WMD \\ \hline
CAAE & 0.458 $\pm$ 0.044 & \textbf{0.498 $\pm$ 0.042} & 0.370 $\pm$ 0.048 & 0.489 $\pm$ 0.043 & 0.496 $\pm$ 0.042 & 0.496 $\pm$ 0.042 \\ \hline
ARAE & 0.337 $\pm$ 0.064 & 0.387 $\pm$ 0.062 & 0.313 $\pm$ 0.065 & 0.419 $\pm$ 0.060 & 0.423 $\pm$ 0.060 & \textbf{0.445 $\pm$ 0.058} \\ \hline
DAR & 0.440 $\pm$ 0.051 & 0.455 $\pm$ 0.050 & 0.379 $\pm$ 0.054 & 0.472 $\pm$ 0.049 & 0.472 $\pm$ 0.049 & \textbf{0.484 $\pm$ 0.048} \\ \hline \hline
Average & 0.412 $\pm$ 0.053 & 0.447 $\pm$ 0.051 & 0.354 $\pm$ 0.056 & 0.460 $\pm$ 0.051 & 0.464 $\pm$ 0.050 & \textbf{0.475 $\pm$ 0.049} \\ \hline
\end{tabular}%
}
\caption{Absolute correlations of content preservation metrics with human scores on texts with style removal.}
\label{table:cp_corr_style_removal}
\end{table*}

\begin{table*}[ht!]
\centering
\resizebox{\textwidth}{!}{%
\begin{tabular}{|l|c|c|c|c|c|c|}
\hline
Model & BLEU & METEOR & Embed Average & Greedy Match & Vector Extrema & WMD \\ \hline
CAAE & 0.488 $\pm$ 0.043 & 0.517 $\pm$ 0.041 & 0.356 $\pm$ 0.049 & 0.490 $\pm$ 0.043 & 0.496 $\pm$ 0.042 & \textbf{0.517 $\pm$ 0.041} \\ \hline
ARAE & 0.356 $\pm$ 0.063 & 0.374 $\pm$ 0.062 & 0.302 $\pm$ 0.066 & 0.405 $\pm$ 0.061 & 0.422 $\pm$ 0.060 & \textbf{0.457 $\pm$ 0.057} \\ \hline
DAR & 0.444 $\pm$ 0.050 & 0.454 $\pm$ 0.050 & 0.370 $\pm$ 0.054 & 0.450 $\pm$ 0.050 & 0.473 $\pm$ 0.049 & \textbf{0.475 $\pm$ 0.049} \\ \hline \hline
Average & 0.429 $\pm$ 0.052 & 0.448 $\pm$ 0.051 & 0.343 $\pm$ 0.056 & 0.448 $\pm$ 0.051 & 0.464 $\pm$ 0.050 & \textbf{0.483 $\pm$ 0.049} \\ \hline
\end{tabular}%
}
\caption{Absolute correlations of content preservation metrics with human scores on texts with style masking.}
\label{table:cp_corr_style_masking}
\end{table*}

We use Fleiss' kappa $\kappa$ of inter-rater reliability \cite[see formula in][]{fleiss73} to identify the more effective human scoring task for different aspects of interest. The kappa metric is often levied in a relative fashion, as there are no universally accepted thresholds for agreements that are slight, fair, moderate, etc. For comprehensive experimentation, we compare kappas over the outputs of each style transfer model. The kappa score for ratings of content preservation based on style-masked texts is $0.297$. Given the kappa score of $0.173$ for unmasked texts, style masking is a more reliable approach towards human evaluation for content preservation (Table \ref{cp_kappas}).

For style transfer intensity, kappas for relative scoring do not show improvement over the previously used approach of absolute scoring of $x'$. However, we observe the opposite for the aspect of naturalness. Kappas for relative naturalness scoring tasks exceed those of the absolute scoring ones (Table \ref{fleiss_nat_full}). Despite the two types of tasks having different numbers of categories (2 vs 5), we can compare them by using a threshold $\tau$ to bin the absolute score for each text into a ``natural" group ($x^\prime$ is considered to be more natural than $x$) or ``unnatural" one (vice versa), like in relative scoring. For example, $\tau=2$ places texts with absolute scores greater than or equal to 2 into the natural group. Judgments for relative tasks yield greater inter-rater reliability than those of absolute tasks across multiple thresholds ($\tau \in \{2,3\}$). This suggests that the relative scoring paradigm is preferable in human evaluations of naturalness.

\subsection{Automated Evaluation}\label{results:ae_cp}
Per aspect of interest, we compute Pearson correlations between scores from the existing metric and human judgments. (As there were three raters for any given scoring task, we take the average of their scores.) We do the same for our proposed metrics to identify which metric is more reliable for automated evaluation of a given aspect.

For style transfer intensity, across both the fastText and textcnn classifiers, our proposed direction-corrected Earth Mover's Distance metric has higher correlation with human scores than the past approach of target style scoring (Table~\ref{sti-corr-full}).

For content preservation, METEOR, shown to have higher correlation with human judgments than BLEU for machine translation \cite{meteor05}, shows the same relationship for style transfer. However, across various text modification settings, WMD generally shows the strongest correlation with human scores (Tables \ref{table:cp_corr_style_removal} and \ref{table:cp_corr_style_masking}). Because WMD is lower when texts are more similar, it is anti-correlated with human scores. We take absolute correlations to facilitate comparison with other content preservation metrics. With respect to text modification, style masking may be more suitable as it, on average for WMD, exhibits a higher correlation with human judgments.

\begin{table}[]
\centering
\resizebox{\columnwidth}{!}{
\begin{tabular}{|c|c|l|l|l|c|l|l|l|}
\hline
\multirow{2}{*}{Model}                & \multicolumn{2}{c|}{Unigram Adv. Clf.}                         & \multicolumn{2}{c|}{Neural Adv. Clf.}  \\ \cline{2-5} 
                                    & Absolute              & \multicolumn{1}{c|}{Relative} & \multicolumn{1}{c|}{Absolute} & \multicolumn{1}{c|}{Relative}  \\ \hline \hline
\multicolumn{1}{|l|}{CAAE} & \multicolumn{1}{c|}{56.07} & \multicolumn{1}{c|}{\textbf{64.51}}                             & \multicolumn{1}{c|}{57.38}                               & \multicolumn{1}{c|}{\textbf{67.87}}  \\ \hline
\multicolumn{1}{|l|}{ARAE}  & \multicolumn{1}{c|}{49.45} & \multicolumn{1}{c|}{\textbf{66.67}}                              &    \multicolumn{1}{c|}{50.68}                           & \multicolumn{1}{c|}{\textbf{67.90}}     \\ \hline
\multicolumn{1}{|l|}{DAR}  & \multicolumn{1}{c|}{65.16} & \multicolumn{1}{c|}{\textbf{65.57}}                           & \multicolumn{1}{c|}{61.07}                            & \multicolumn{1}{c|}{\textbf{62.30}}     \\ \hline \hline
\multicolumn{1}{|l|}{Average}  & \multicolumn{1}{c|}{56.89} & \multicolumn{1}{c|}{\textbf{65.58}}                           & \multicolumn{1}{c|}{56.38}                            & \multicolumn{1}{c|}{\textbf{66.02}}     \\ \hline
\end{tabular}
}
\caption{Percent agreement between adversarial classifiers and human scores on the naturalness of texts.}
\label{agreement-nt-full}
\end{table}

\begin{figure*}[t]
  \center
  \includegraphics[width=\columnwidth]{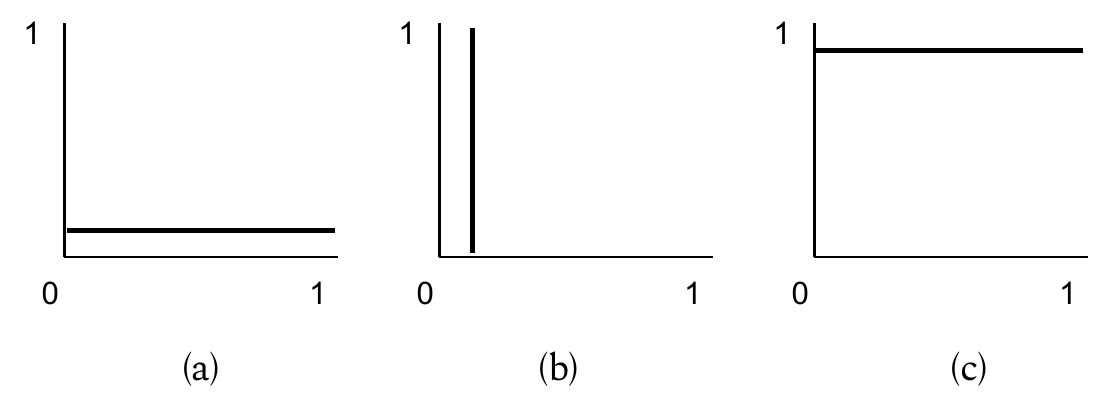}
  \caption{Extreme tradeoff plots, with style transfer intensity on the x-axis and content preservation on the y-axis.}
  \label{fig:extreme_tradeoffs}
\end{figure*}

\begin{figure*}[t] %[htp]
  \centering
  \begin{subfigure}[b]{\columnwidth}
    \includegraphics[width=\columnwidth]{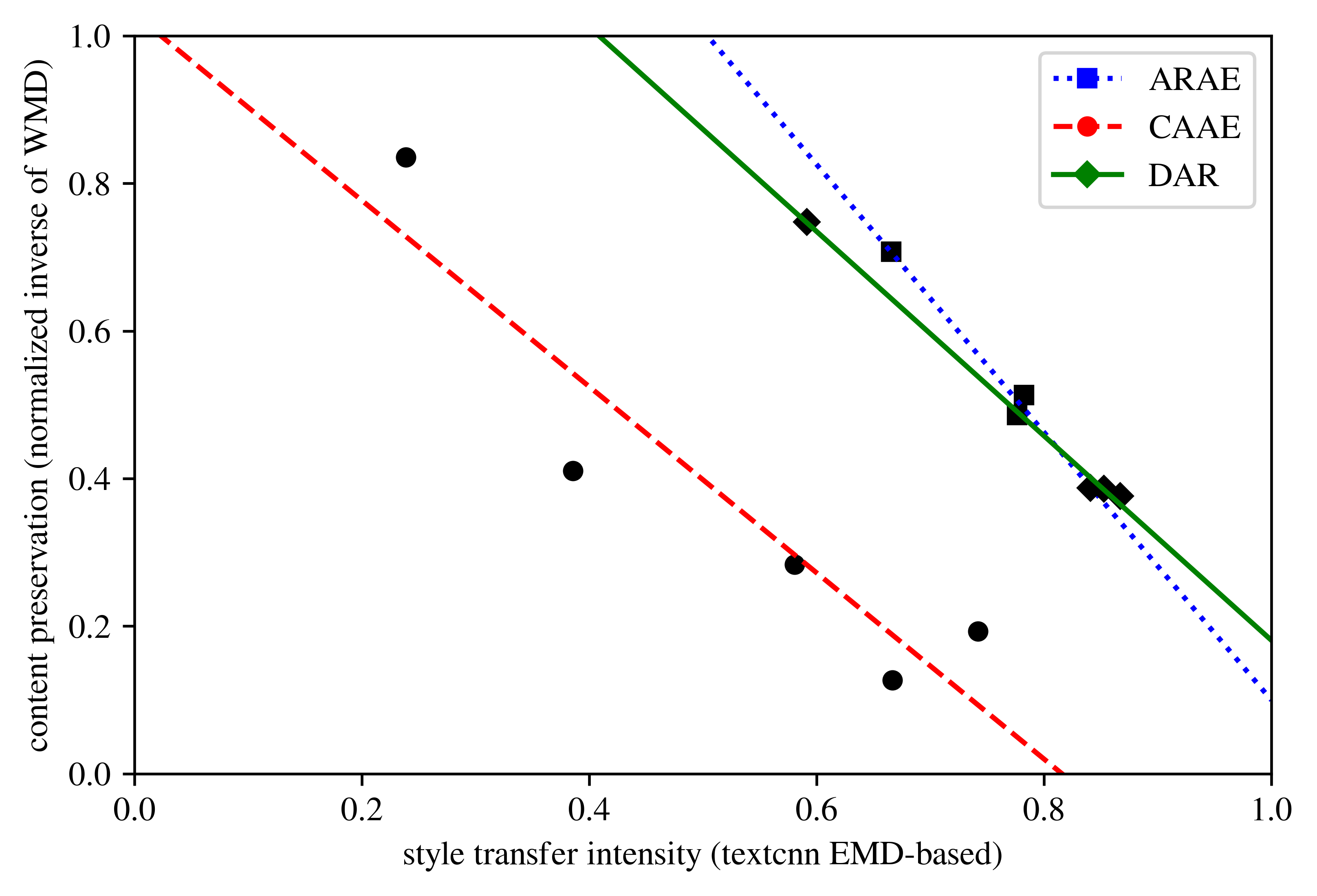}
	\caption{Content vs. Style Tradeoffs}
  \end{subfigure}
  \begin{subfigure}[b]{\columnwidth}
	\includegraphics[width=\columnwidth]{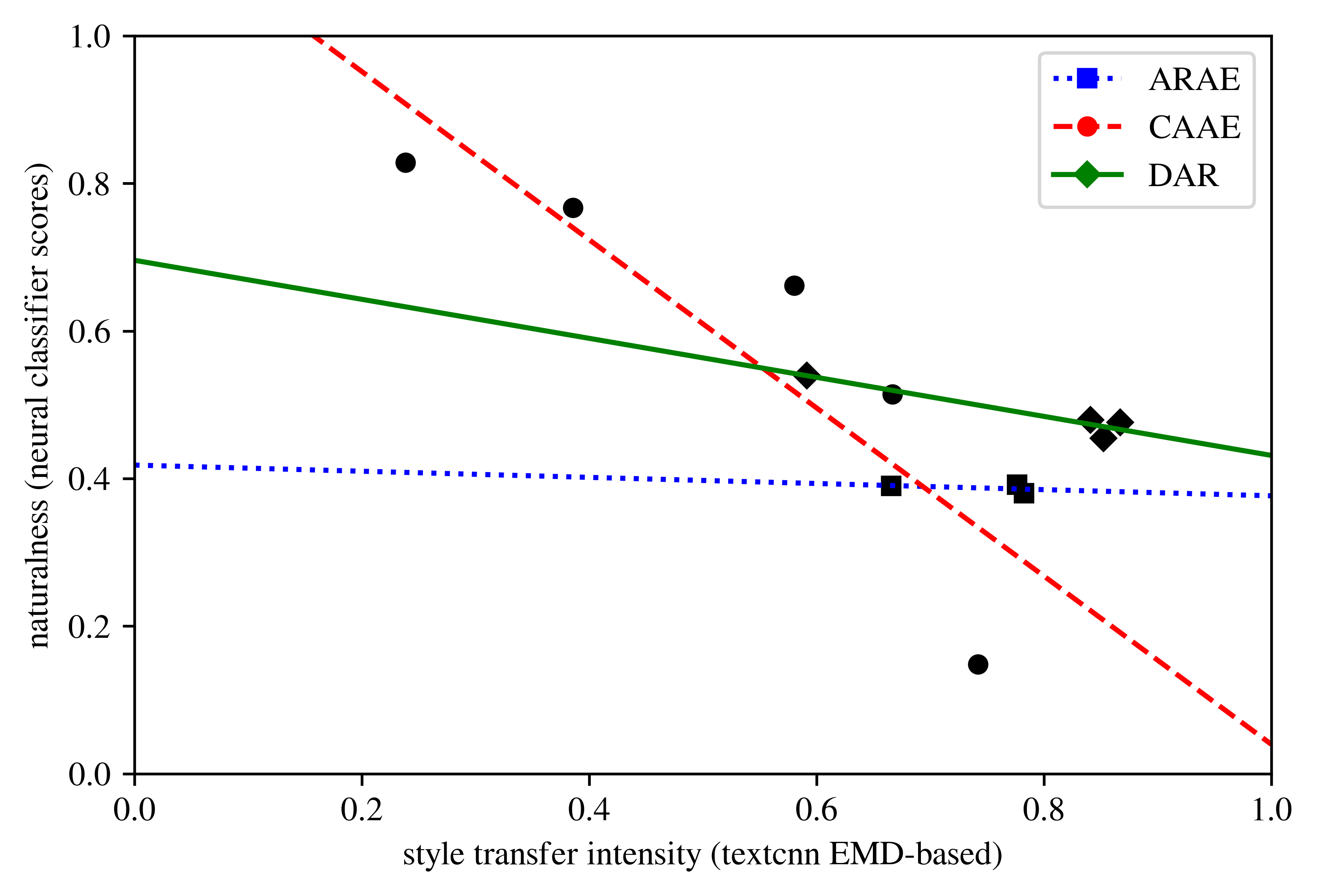}
    \caption{Naturalness vs. Style Tradeoffs}
  \end{subfigure}
  \caption{Tradeoffs between aspects of evaluation, using metrics most strongly correlated with human scores.}
  \label{tradeoff}
\end{figure*} 

For naturalness, both unigram and neural classifiers exhibit greater agreement on which texts are considered more natural with the humans given relative scoring tasks than with those given absolute scoring tasks (Table \ref{agreement-nt-full}), although the neural classifier achieves higher agreements on average. We also confirm that sentence-level perplexity is not an appropriate metric. It exhibits no significant correlation with human scores ($\alpha = 0.05$). These results suggest that adversarial classifiers can be useful for automating measurement of naturalness.

\subsection{Aspect Tradeoffs}

Previous work has compared models with respect to a single aspect of interest at a time, but has only, to a limited degree, considered how relationships between multiple aspects influence these comparisons. In particular, concurrent work by \cite{StanfordDAR18} examines tradeoff plots, but focuses primarily on variants of its own model, while including only a single point on the plots of style transfer models from other papers. For a comprehensive comparison, it is ideal to have plots for all models.

It is helpful to first understand the tradeoff space. For example, we define extreme cases for style transfer intensity and content preservation, where we assume measurement of the latter ignores stylistic content. Consider two classes of suboptimal models. One class produces outputs with a wide range of style transfer intensity, but poor content preservation (Figure \ref{fig:extreme_tradeoffs}a). The other class of models produces outputs with low style transfer intensity, but a wide range of content preservation (Figure \ref{fig:extreme_tradeoffs}b).

This is in contrast to a model that yields a wide range of style transfer intensity and consistently high content preservation (Figure \ref{fig:extreme_tradeoffs}c). If we take that to be an ideal model for a sentiment dataset, we can interpret models with better performance to be the ones whose tradeoff plots are closer to that of the ideal model and farther from those of the suboptimal ones. The plot for an ideal model will likely vary by dataset, especially because the tradeoff between content preservation and style transfer intensity depends on the level of distinction between style words and content words of the dataset.

With this interpretation of the tradeoff space, we construct a plot for each style transfer model (Figure~\ref{tradeoff}), where each point represents a different hyperparameter setting for training (Section \ref{model_details}). We collect scores based on the automated metrics most strongly correlated with human judgment: direction-corrected EMD for style transfer intensity, WMD for content preservation, and percent of output texts marked by an adversarial classifier as more natural than input texts. Because WMD scores are lower when texts are more similar, we instead take the normalized inverses of the scores to represent the degree of content preservation.

Across all models, there is a trend of reduction in content preservation and naturalness as style transfer intensity increases. Without the plots, one might conclude that ARAE and DAR perform substantially differently, especially if hyperparameters are chosen such that ARAE achieves the leftmost point on its plot and DAR achieves the rightmost point on its plot. With the plots, at least for the set of hyperparameters considered, it is evident that they perform comparably (Figure~\ref{tradeoff}a) and do not exhibit the same level of decrease in naturalness as CAAE (Figure~\ref{tradeoff}b).

\section{Discussion} 
Previous work on style transfer models used a variety of evaluation methods (Table~\ref{relatedwork_summary}), making it difficult to meaningfully compare results across papers. Moreover, it is not clear from existing research how exactly to define particular aspects of interest, or which methods (whether human or automated) are most suitable for evaluating and comparing different style transfer models.

To address these issues, we specified key aspects of interest (style transfer intensity, content preservation, and naturalness) and showed how to obtain more reliable measures of them from human evaluation than in previous work. Our proposed automated metrics (direction-corrected EMD, WMD on style-masked texts, and adversarial classification) exhibited stronger correlations with human scores than existing automated metrics on a binary sentiment dataset. While human evaluation may still be useful in future research, automation facilitates evaluation when it is infeasible to collect human scores due to prohibitive cost or limited time.

\subsection{Human Evaluation}
For style transfer intensity, the relative scoring task (rating the degree of stylistic difference between $x$ and $x^\prime$) did not have greater rater reliability than the previously used task of rating output texts on an absolute scale. This is likely due to task complexity or rater uncertainty, which motivates the need for further exploration of task design for this particular aspect of interest. 

For content preservation, our form of human evaluation operates on texts whose style words are masked out, unlike the previous approach (no masking). Our approach addresses the unintentional variable of rater-dependent style identification that could lead to noisy, less reliable ratings.

Identification and masking of words was made possible with a style lexicon. We automatically constructed the lexicon in a way that can be done for any style dataset, as long as style labels are available (Section~\ref{methods_style_lexicon}). We acknowledge a tradeoff between filling the lexicon with more style words and being conservative in order to avoid capturing content words. We justify taking a more conservative approach as content words are naturally critical to evaluations of content preservation.

For naturalness, we introduced a paradigm of relative scoring that uses both the output and input texts. This achieved a higher inter-rater reliability than did absolute scoring, the previous approach. 

\subsection{Automated Evaluation}
For style transfer intensity, we proposed using a metric with EMD as the basis to acknowledge the spectrum of styles that can appear in outputs and to handle both binary and non-binary datasets. The metric also accounts for direction by penalizing scores in the cases where the style distribution of the output text explicitly moves away from the target style. Previous work used external classifiers, whose style distributions for $x$ and $x^\prime$ can be used to calculate direction-corrected EMD, making it a simple addition to the evaluation workflow.

For content preservation, WMD (based on EMD) works in a similar fashion, but with word embeddings of $x$ and of $x^\prime$. BLEU, used widely in previous work, may yield weaker correlations with human judgment in comparison as it was designed to have multiple reference texts per candidate text \cite{Papineni02}. Several reference texts, which are more common in machine translation tasks, increase the chance of $n$-gram (such as $n\geq3$) overlap with the candidate. In the style transfer setting, however, the only reference text for $x^\prime$ is $x$. Having a single reference text reduces the likelihood of overlap and the overall effectiveness of BLEU.}

For naturalness, strong agreement of adversarial classifiers with relative scores assigned by humans suggest that classifiers are suitable for automated evaluation. One might assume input texts would almost always be rated as more natural by both humans and classifiers, biasing the agreement. This is not the case, as we justify our rating scheme with evidence of outputs being rated as more natural across several models (Figure~\ref{tradeoff}b). Output texts classified as more natural indicate some success for a style transfer model, as it can produce texts with a quality like that of human-generated inputs, which are, by definition, natural.

Finally, with aspect tradeoff plots constructed using scores from the automated metrics, we can directly compare models with respect to multiple aspects simultaneously. Points of intersection, or near intersection, for different models signify that they, at the hyperparameters that yielded those points, can achieve similar results for various aspects. These parameters can be useful for understanding the impact of decisions made during model design and optimization phases.

\subsection{Future Research}
As we confirmed, sentence-level perplexity of output $x^\prime$ is not meaningful by itself for the automated evaluation of naturalness. The idea of using both $x$ and $x^\prime$, akin to how we train automated classifiers of naturalness (Section \ref{methods_autometrics}), can be extended to construct a perplexity-based metric that also takes into account the perplexity of input $x$.

Another avenue for future work could be evaluating on datasets with a different style or number of style classes. It is worth studying the distinction between style words and content words in the vocabulary of each such dataset. Given the definition of style transfer and its simplifying assumption in Section \ref{intro}, it would be reasonable to expect naturally low content preservation scores for any given style transfer model operating on datasets with less distinction, such as those of formality. This is not so much an issue as it is a dataset-specific trend that can be visualized in corresponding tradeoff plots, which would provide a holistic evaluation of model performance. In any case, results from inter-rater reliability and correlation testing on these additional datasets would overall enable more consistent evaluation practices and further progress in style transfer research.

\section*{Acknowledgments}
We would like to thank Juncen Li, Tianxiao Shen, and Junbo (Jake) Zhao for guidance in the use of their respective style transfer models. These models serve as markers of major progress in the area of style transfer research, without which this work would not have been possible.

\bibliography{naaclhlt2019}
\bibliographystyle{acl_natbib}

\end{document}